\documentclass{article}

\usepackage{arxiv}

\usepackage[utf8]{inputenc} % allow utf-8 input
\usepackage[T1]{fontenc}    % use 8-bit T1 fonts
\usepackage{hyperref}       % hyperlinks
\usepackage{url}            % simple URL typesetting
\usepackage{booktabs}       % professional-quality tables
\usepackage{amsfonts}       % blackboard math symbols
\usepackage{nicefrac}       % compact symbols for 1/2, etc.
\usepackage{microtype}      % microtypography
\usepackage{lipsum}
\usepackage{graphicx}
\usepackage[export]{adjustbox}
\usepackage{multirow}
\usepackage{breqn}

\title{Estimation of Continuous Blood Pressure from PPG via a Federated Learning Approach.}

\author{
  \makebox[.4\linewidth]{Eoin Brophy} \\
  Infant Research Centre \& School of Computing\\
  Dublin City University\\
  Ireland \\
  \texttt{eoin.brophy7@mail.dcu.ie} \\
  \And
 \makebox[.4\linewidth]{Maarten De Vos} \\
  Department of Electrical Engineering\\
  KU Leuven\\
  Belgium \\
  \And
 \makebox[.4\linewidth]{Geraldine Boylan} \\
  Infant Research Centre\\
  Cork University Maternity Hospital\\
  Ireland\\
   \And
 \makebox[.4\linewidth]{Tom{\'a}s Ward} \\
  Insight SFI Research Centre\\
  Dublin City University\\
  Ireland \\

}

\begin{document}
\maketitle

\begin{abstract}
Ischemic heart disease is the highest cause of mortality globally each year. This not only puts a massive strain on the lives of those affected but also on the public healthcare systems. To understand the dynamics of the healthy and unhealthy heart doctors commonly use electrocardiogram (ECG) and blood pressure (BP) readings. These methods are often quite invasive, in particular when continuous arterial blood pressure (ABP) readings are taken and not to mention very costly. Using machine learning methods we seek to develop a framework that is capable of inferring ABP from a single optical photoplethysmogram (PPG) sensor alone. We train our framework across distributed models and data sources to mimic a large-scale distributed collaborative learning experiment that could be implemented across low-cost wearables. Our time series-to-time series generative adversarial network (T2TGAN) is capable of high-quality continuous ABP generation from a PPG signal with a mean error of $2.54  mmHg$ and a standard deviation of $23.7  mmHg$ when estimating mean arterial pressure on a previously unseen, noisy, independent dataset. To our knowledge, this framework is the first example of a GAN capable of continuous ABP generation from an input PPG signal that also uses a federated learning methodology.

\end{abstract}

% keywords can be removed

\section{Introduction}
Chronic heart disease was the number one cause of death from 2000-2019 according to the World Health Organisation (WHO) and was responsible for 16\% of the total worldwide deaths in 2019 \cite{WHO_2020}. Heart disease has also shown the largest increase in deaths during this period. Obtaining unobtrusive continuous measurements of the cardiac state has proven very difficult. The most commonly used indicator for a measurement of the state of the heart is blood pressure (BP) which is often gathered using a sphygmomanometer cuff, finapres, or an arterial catheter.  Sphygmomanometers provide spot-measurements for BP over a very short time interval and arterial catheters are an extremely invasive method of continuous BP measurement. The finapres is an alternative for continuous and unobtrusive BP measurement however the size, shape, and price of these devices means that they have not been commoditised for individuals seeking continuous home BP measurement devices. Regular monitoring of BP can prove vital for people suffering from cardiovascular diseases (CVDs) who are already vulnerable to BP fluctuations.

Methods for non-invasively measuring continuous arterial blood pressure (ABP) have been explored that use other physiological signals to infer ABP. One example uses the pulse transit time (PTT) that is the time interval taken for a pulse wave to travel between two arterial sites. PTT varies inversely to BP changes and has been demonstrated to be a valid and accepted measure of BP \cite{padilla_2006, wong_evaluation_2009}.  As PTT is defined as the difference in the R-wave interval of an electrocardiogram (ECG) signal. This information should also be available from a photoplethysmography (PPG) signal.

PPG is an optical technique that requires a single sensor and has become commoditised in the past number of years such that it is included in most wearables and other medical devices. It works by shining a light-emitting diode (LED) into the microvascular tissue and measuring the amount of light reflected/transmitted and absorbed via a photo-sensor and detects the volume changes of blood over the cardiac cycle. The output from this sensor is then conditioned so a valid heart rate can be determined. Having a continuous heart rate measurement means we can extract a PTT measurement also which means that providing further analysis we can extract meaningful ABP measurements using a PPG sensor alone. 

Our work described here is part of a larger-scoped effort to develop readily-deployable artificial intelligence (AI) systems that can be easily interpreted by non-expert consumers and downstream end-users. Capitalising on recent advancements in machine learning has the potential to simplify wearables devices, allowing for a reduction in power requirements and subsequently, lower-cost devices as our previous work also aims to achieve \cite{brophy2020optimised}.

In this paper, we present our novel framework for implementing continuous ABP measurement using a PPG sensor alone. Our methods seek to use proven cutting-edge machine learning techniques to capture the characteristics that correlate and link continuous PPG to continuous ABP measurements. For the first time, we demonstrate a decentralised learning approach to continuous ABP measurement that is capable of real-world implementation on a large scale and does not compromise patient privacy. This novel approach yields not only a more power-efficient learning framework thus advancing the development of simpler, more cost-effective wearables but does so without compromising the accuracy of ABP measurements and patient privacy.

\section{Related Work}
Many works in the past have focused on estimating BP from correlating features available in ECG and PPG \cite{padilla_2006, wong_evaluation_2009, teng_continuous_2003, Lin_2021}. These works have demonstrated high-quality results in the estimation of ABP but require domain expertise to process the available PPG signal to acquire the blood pressure estimation. A growing number of machine learning methods are being developed to remove the dependency on signal processing experts allowing for readily-deployable AI systems. Our framework follows suit in automating the signal extraction process, making handcrafted feature selection obsolete.

Slapni\v{c}ar et al. implemented a spectro-temporal deep neural network (DNN) to model the dependencies that exist between PPG and BP waveforms. The authors used a PPG signal along with its first and second derivatives and determined the network successful at modelling the dependent characteristics of BP \cite{slapnicar_blood_2019}. El Hajj and Kyriacou implement recurrent neural networks (RNNs) for estimation of BP from PPG only \cite{Hajj_2020}. Other works develop a statistical feature extraction and selection process followed by a regression-based predictive model, all of which achieve high-quality BP estimation results from PPG data only \cite{Nath_2020}. Feature-free methods of BP estimation have also been completed previously through the use of a deep-learning-based prediction techniques with good results \cite{Schlesinger_2020, Panwar_2020}. However, these methods discussed thus far deal with BP prediction in discrete intervals, we build on this through the generation of continuous BP waveforms and BP prediction.

Another feature-free deep learning method of non-invasive continuous blood pressure modelling was introduced by Ibtehaz and Rahman \cite{ibtehaz_ppg2abp_2020}. The authors presented their PPG2ABP method that utilises a deep-supervised U-Net model that consists of an encoder and decoder network adopted for regression. In this configuration, their model can predict a complete continuous BP waveform from a PPG waveform. To build on these works our framework implements a Long short-term memory (LSTM) convolutional neural network (CNN) (LSTM-CNN) GAN model that is capable of generating continuous BP from a given PPG signal. Not only is our model capable of PPG2ABP but also ABP2PPG, this enables our model to infer one physiological time series waveform from another. Our model can map any given time series signal to another but for the transform to make sense the signals should be correlated in some way.

Smartwatches have become pervasive in recent years but are still technically lacking in terms of sensors available to the end-users. Mena et al. developed a mobile system for non-invasive, continuous BP monitoring \cite{Mena_2020}. Their system actively collected and transmitted PPG readings from a wrist-mounted sensor to a smartphone where BP estimation is computed with machine learning algorithms. Our framework complements and builds on this approach in employing a decentralised multi-user learning framework enabling more accurate predictive models and in-turn faster patient diagnoses.

PPG has become a staple sensor in wearables and the main means of measuring the heart rate of end-users yet from a medical perspective ECG is the proven and more information-rich signal as a measure of the cardiac state. In addressing this issue Sarkar and Etemad \cite{sarkar_cardiogan_2020} present their model CardioGAN that employs an adversarial training method to map PPG to ECG signals. CardioGAN utilises both time and frequency-domain features of the PPG to generate reliable 4-second long ECG signals. Our method implements a time-domain only discriminator to reduce an individual model's overhead and is capable of generating 10-second long PPG and ABP waveforms from one another. We also take into account the personal data preserving methods and demonstrate real-world applications of such models. 

Addressing data sharing and privacy issues we adopt a federated learning approach that was originally introduced by Google's AI Blog \cite{Google_FL_2017} as a means to collaborate machine learning across mobile devices without the need to store data in a centralised repository. It enables remote devices to learn collaboratively with a shared global-model while keeping their training data on individual devices. The individual client-models train locally and send their weights to be aggregated on the global-model which can then be passed back as updated training weights for the client-models where training can continue. This entire process is known as a communication round. Rasouli et al. \cite{rasouli2020fedgan} presented one of the first examples of implementing this training process as part of a GAN on image and energy data. We implement a similar training strategy for our framework to serve as proof of concept for distributed training across smartwatches to build a model such as the one presented in our paper.

\section{Methodology}
\begin{figure}[ht]
    \centering
    \includegraphics[width=0.85\textwidth]{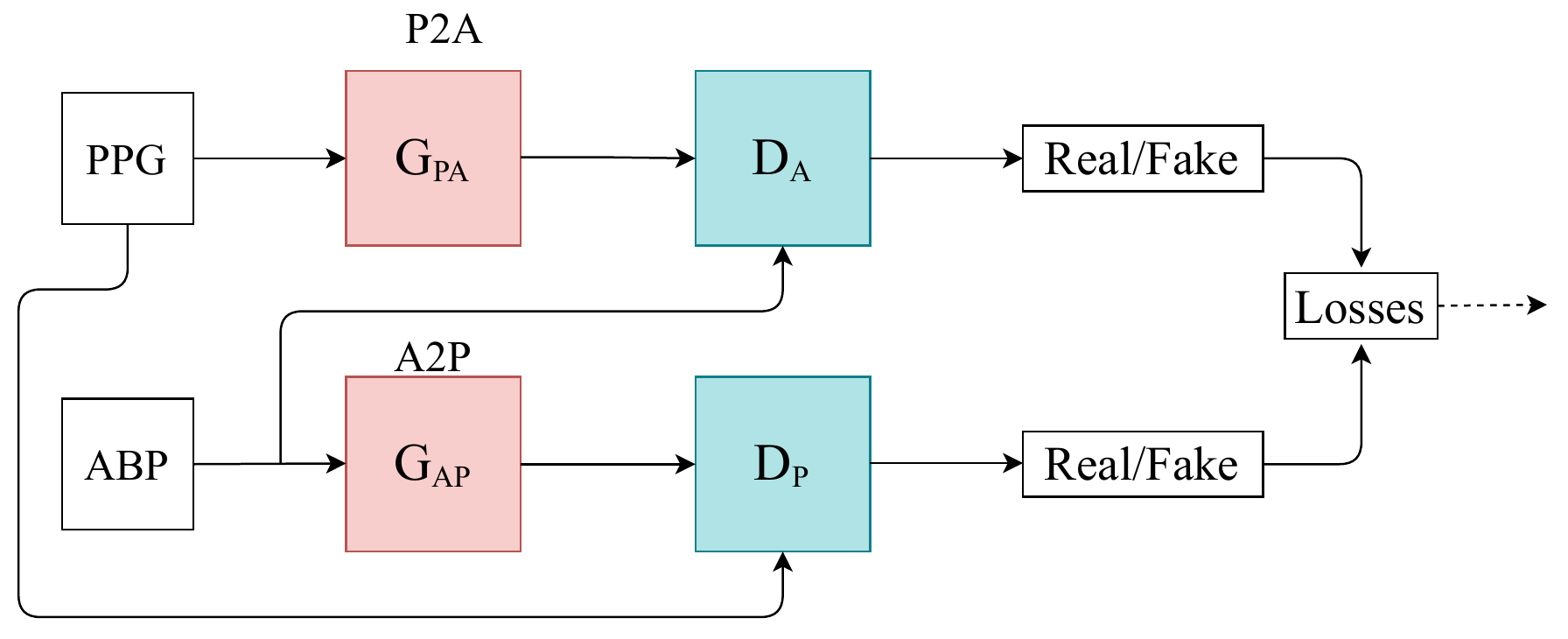}
    \caption{Architecture of the T2T-GAN. P2A represents the generator transform function from PPG to ABP. Conversely, A2P represents ABP to PPG.}
    \label{fig:T2TGAN}
\end{figure}

We design a Time series to Time series Generative Adversarial Network (T2T-GAN) (Figure \ref{fig:T2TGAN}), based on the popular CycleGAN that is capable of unpaired image-to-image translation \cite{zhu2020CycleGAN}. The T2T-GAN can translate from one time series modality to another using cycle-consistency losses. More specifically, we implement the T2T-GAN for capturing the complex characteristic relationship between ABP and PPG and train this model to translate a PPG measurement into an accurate continuous ABP measurement. In the interest of data privacy and protection and real-world implementation, we opt for a decentralised learning approach here and implement federated learning. Having one central aggregate model and many decentralised models we can implement our framework without ever needing to handle individuals' personal sensitive data. Comprehensive details of our method can be found in the section that follows.

\subsection{Computing Platform}
The experiments for this project were run on an Nvidia Titan Xp with PyTorch and Google Colaboratory in interest for making the project readily deployable. The code for these experiments are available online\footnote{GitHub Repository: https://github.com/Brophy-E/T2TGAN}.

\subsection{Dataset}
Two open-source datasets were used in this experiment. The first dataset "Cuff-Less Blood Pressure Estimation" is freely available on both Kaggle and UCI Machine Learning Repository. It contains preprocessed and validated ECG (electrocardiogram from channel II), PPG (fingertip) and ABP (invasive arterial blood pressure (mmHg)) signals all sampled at 125 Hz \cite{kachuee_cuff-less_2015, kachuee_cuffless_2017}. The raw ECG, PPG, and ABP signals were originally collected from PhysioNet \cite{PhysioNet}. This dataset is split into multiple parts and consists of several records, for our work we used the first 5 ($part1.mat - part5.mat$) records and segmented them into 8-second intervals, that yielded 144000 training samples (320 hours) and the last 2 ($part11.mat - part12.mat$) records into 55000 validation samples  (122 hours). However, as there might be more than one record per patient (which is not possible to distinguish) we use a second unrelated dataset to test our framework and observe its generalisability. Therefore we used an $[144000, 2, 1000]$ dimensional vector that constituted the training dataset for our framework.

\begin{figure}[ht]
    \centering
    \includegraphics[width=0.5\textwidth]{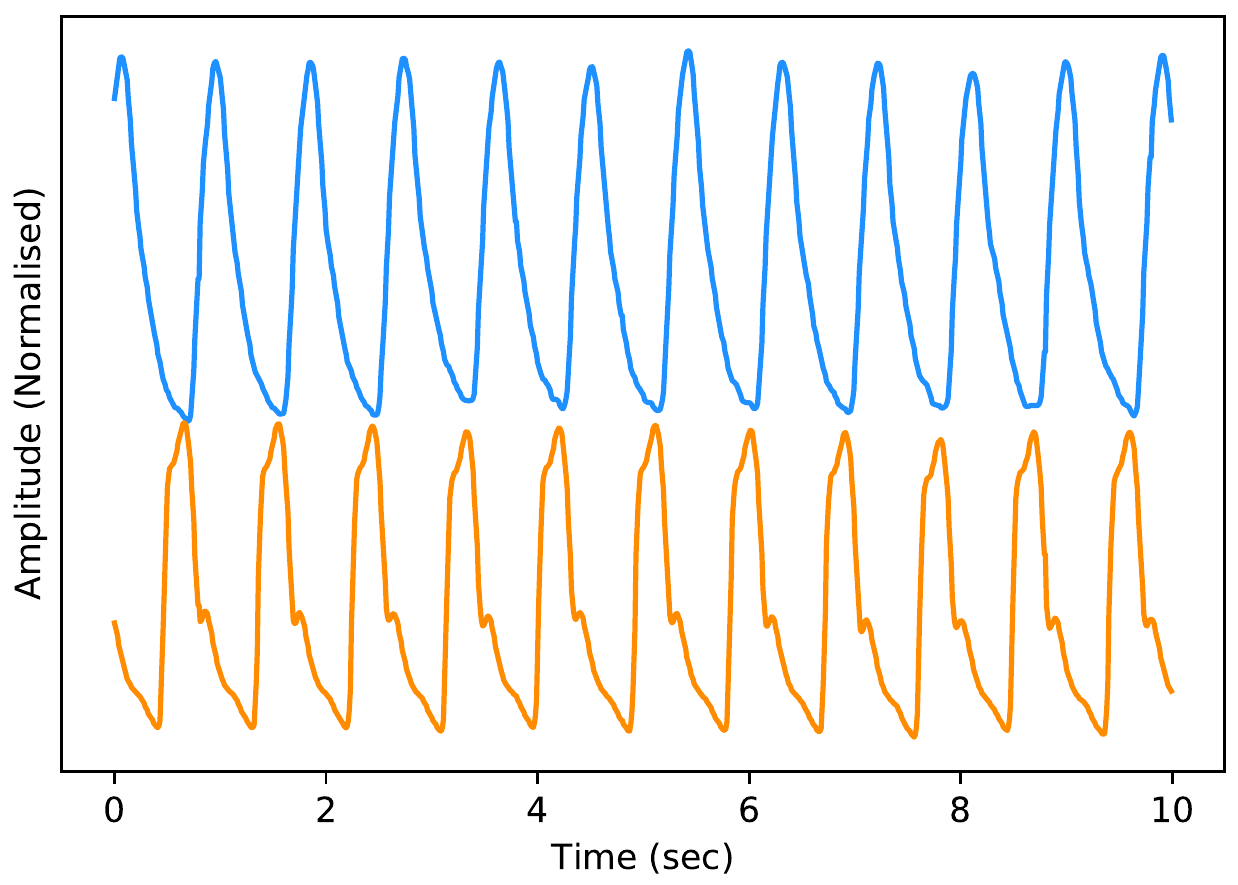}
    \caption{Example of Real PPG (top, blue) and ABP (bottom, orange). The signals are both normalised between 0 and 1 with an artificial offset on the ABP signal for visualisation purposes}
    \label{fig:RealSignals}
\end{figure}

The test dataset "University of Queensland vital signs dataset: development of an accessible repository of anesthesia patient monitoring data for research" \cite{liu_university_2012} provides a multitude of vital sign waveform data recorded from patients undergoing anesthesia at the Royal Adelaide Hospital. The physical state of patients under anesthesia contains marked changes to cardiovascular variables compared to ICU patients which serve to present a further challenge to our framework. We are concerned with the ABP and PPG measurements only from this dataset, these are sampled at 100 Hz. We select only one patient, namely \textit{Case 5} and segment the data into 10-second intervals which yields an $[900,2,1000]$ dimensional vector (150 minutes) that constitutes the test dataset for our framework. We are only concerned with the PPG and ABP signals from these datasets, see Figure \ref{fig:RealSignals} for an example of the real data used in this work.

\subsection{Model}
\begin{figure}[ht]
    \centering
    \includegraphics[width=0.75\textwidth]{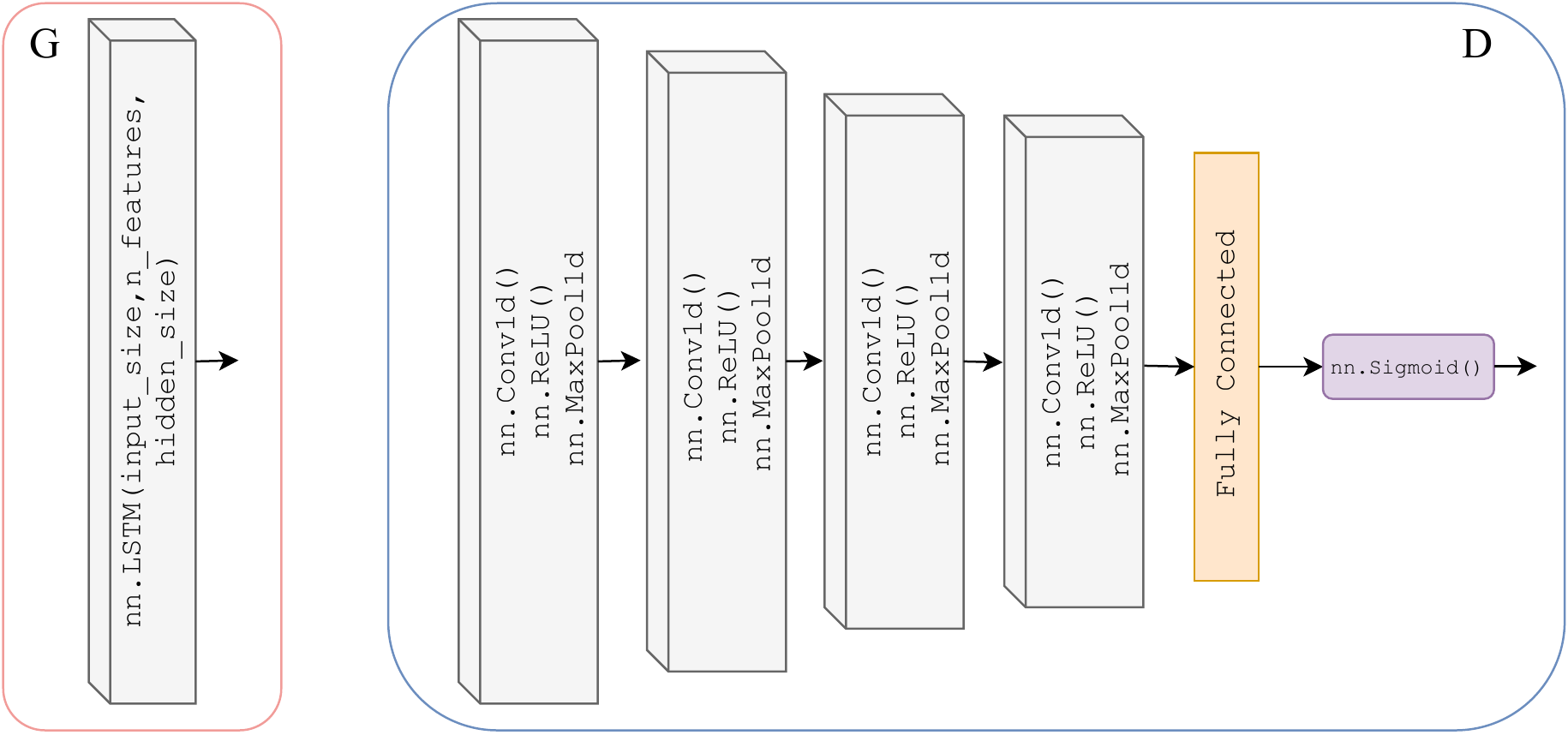}
    \caption{Architecture of Generators $G_{PA}$ and $G_{AP}$ (left) which are two-layer stacked LSTMs with 50 hidden units in each layer and a fully connected layer at the output, with an input size of 1000. Architecture of Discriminators $D_{A}$ and $D_{P}$ (right) which are 4-layer 1-dimensional CNNs (ReLU activation and max pooling functions) with a fully connected layer and sigmoid activation function at the output.}
    \label{fig:LSTM_CNN}
\end{figure}

As previously mentioned, we adopted the learning framework of CycleGAN for time series data to translate from one time series modality to another. Here we will explicitly define the Discriminator and Generator architecture of our T2T-GAN. 
The Generators $G_{PA}$ and $G_{AP}$ are two-layer stacked LSTMs with 50 hidden units in each layer and a fully connected layer at the output, with an input size of 1000, see Figure \ref{fig:LSTM_CNN}. The Discriminators $D_{A}$ and $D_{P}$ are 4-layer 1-dimensional CNN with a fully connected layer and sigmoid activation function at the output, see Figure \ref{fig:LSTM_CNN}.

\subsection{Federated Learning}
\begin{figure}[ht]
    \centering
    \includegraphics[width=0.45\textwidth]{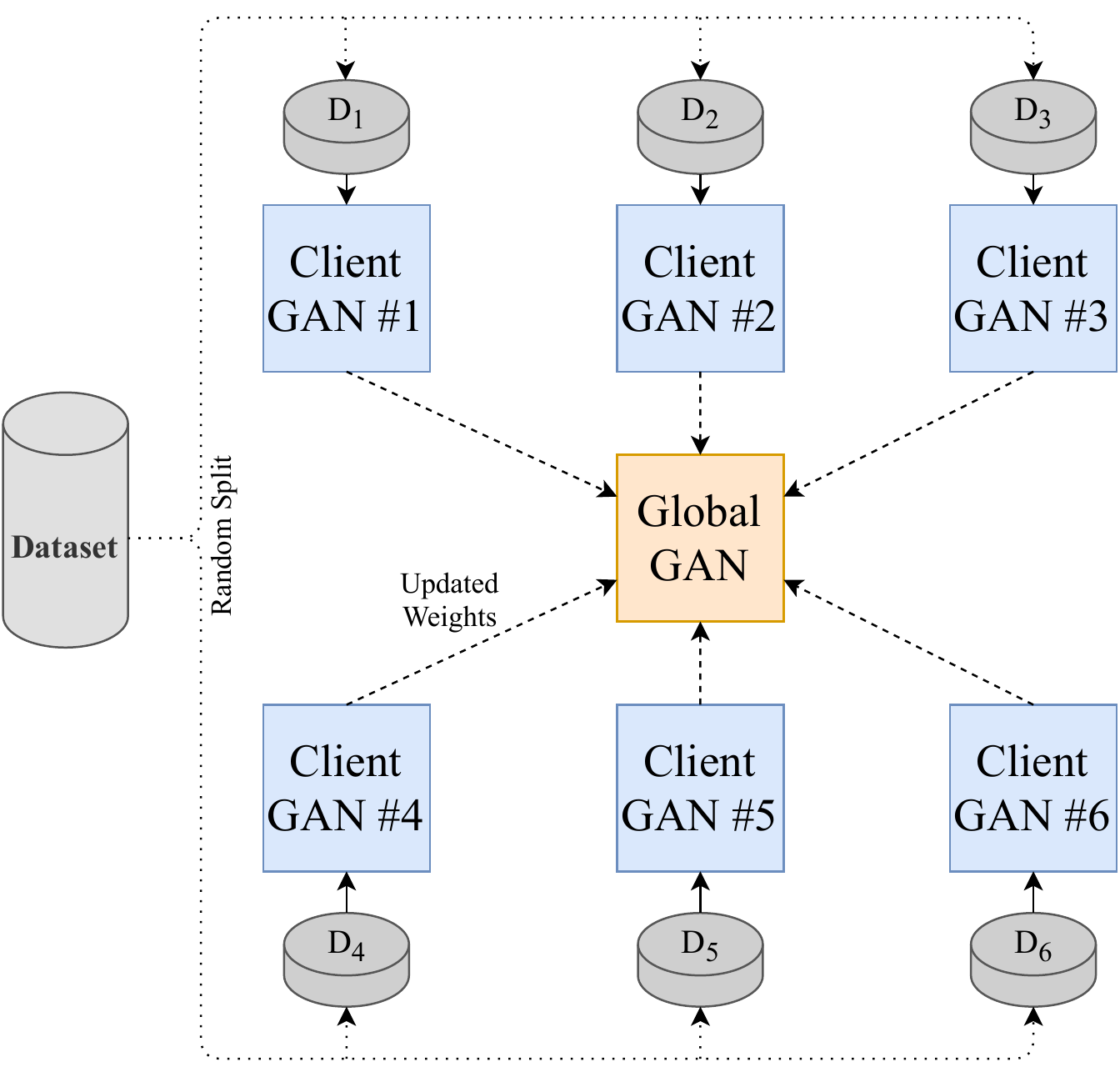}
    \hspace{20mm}
    \includegraphics[width=0.35\textwidth]{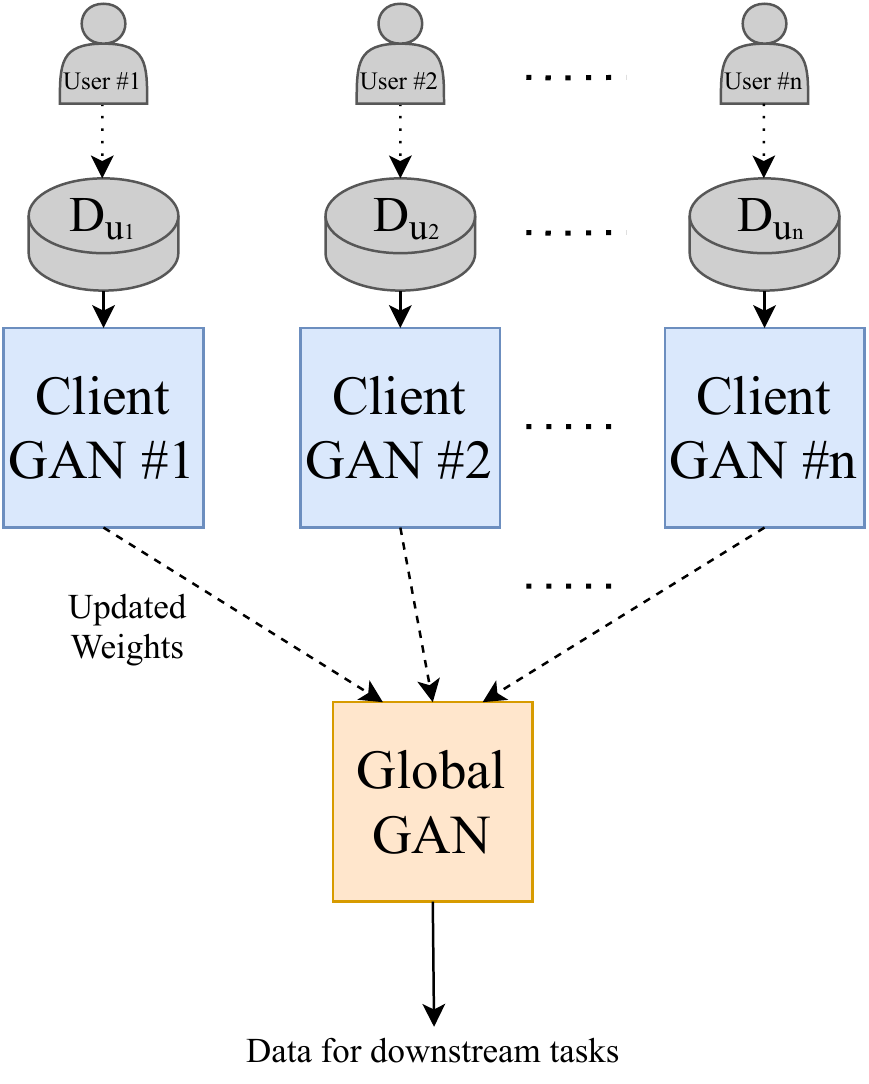}
    \caption{Federated Learning methodology employed in this paper (Left). Each GAN is represented by the model shown previously in Figure \ref{fig:T2TGAN}. Federated Learning methodology that is implemented in the real-world (Right).}
    \label{fig:FedGAN}
\end{figure}

To make the model perform closer to a real-world setting and to prevent data sharing to third-parties we implement the decentralised learning approach of Federated Learning. Our approach is limited to using one central server so to realise this learning method we split our dataset into $N$ (where $N=20$) equally sized random smaller data subsets and train $N$ client-GANs on their own data with no cross-over from their respective subsets. The client-GANs are trained for $e$ (where $e=5$) epochs and their weights are then sent to a global-GAN that aggregates the received weights from the $N$ clients-GANs. This global-GAN can then operate on unseen data or update the client-GANs with the aggregated global weights which eliminate the need for any data centralisation, see Figure \ref{fig:FedGAN} (left) for a visual example of our method. Of course in a real-world training and testing environment, the training data will not come from a centralised repository. The data will instead be generated by the end-users. Consumers will generate their own PPG data from their smartwatch, in this case, that will be used to train a local-model and communicate weights to and from a global-model, see Figure \ref{fig:FedGAN} (right) for a conceptual example.

\subsection{Training}

We chose a total of 20 client-models for training as demonstrated in Figure \ref{fig:FedGAN}. Each used an equal proportion of the dataset. 6 random clients were selected from the total client-models in each communication round to be trained. There were 10 communication rounds, following each round of training on the client-device the aggregation of weights is computed on the global-model. The total number of training rounds on each client was 5, with a batch size of 32. 
The total loss function of our T2T-GAN framework is calculated as:

\begin{dmath}
\mathcal{L}(P2A,A2P,D_{P},D_{A})  = \mathcal{L}_{T2T-GAN}(P2A,D_{A},PPG,ABP) \\
+ \mathcal{L}_{T2T-GAN}(A2P,D_{P},ABP,PPG) \\
+ \lambda_{c} \mathcal{L}_{cyc}(P2A,A2P)
+ \lambda_{i} \mathcal{L}_{identity}(P2A,A2P)
\label{eq:1}
\end{dmath}

where $\mathcal{L}_{cyc}$ and $\mathcal{L}_{identity}$ is the L1-norm and $\mathcal{L}_{T2T-GAN}$ is defined as the mean squared error loss (MSE). $\lambda$ controls the relative importance of the two objectives, $\lambda_{c}$ and $\lambda_{i}$ were chosen as 10 and 5 respectively.

\subsection{Evaluation}\label{sec:eval}
To successfully evaluate our model we examine the mean arterial pressure (MAP) of generated samples. Using a completely independent test dataset from the training dataset grants us the freedom to implement a leave-one-out strategy and see how well our model generalises to other ABP-PPG datasets. We take the PPG measurements from the test dataset and pass them through our trained global deterministic function; P2A. This function converts our PPG signal to a corresponding ABP signal and we then calculate the MAP from the generated signal and compare it with the true MAP measurements from the real ABP signal. MAP is considered a better indicator of perfusion to vital organs than systolic blood pressure (SBP) \cite{MAP_2011}. It is important to note that we can retrieve the systolic and diastolic blood pressure (DBP) from the P2A signal which we use to calculate the MAP (\ref{eq:2-MAP}) rather than simply returning the mean of the continuous signal segment. We also present the Bland-Altman (BA) plots of the MAP error \cite{Giavarina_2015} that allow us to determine to what degree the generated ABP is a good substitute for the real ABP. The Association for the Advancement of Medical Instrumentation (AAMI) standard requires BP measuring methods to have a mean error ($\mu$) and standard deviation ($\sigma$) of less than $5 mmHg$ and $8 mmHg$ respectively \cite{AAMI_1987}. Following this, we then select the entire 150-minute period of the test-data and perform calibration on this data for the first one-minute period only. This calibration is designed to remove user bias and provide more accurate results while mimicking a continuous BP measurement test that can be performed clinically. Bland-Altman plots are provided for the calibrated and uncalibrated measurements.

\begin{dmath}
MAP = [SBP + (2*DBP)] / 3
\label{eq:2-MAP}
\end{dmath}

In the interest of providing a comprehensive evaluation of our T2TGAN we implement the dynamic time warping (DTW), root-mean-squared error (RMSE) and Pearson Correlation Coefficient (PCC) algorithms as distance and similarity measures between the real and generated time series BP sequences for both the federated and un-federated approaches. These metrics allow us to quantify the similarities in structure of the blood pressure waveforms. This is implemented for the entire test dataset (150 minutes, 900 samples at 10 sec/sample) and a random sample of the validation dataset of equal size.

\section{Results}
As stated previously in Section \ref{sec:eval} we evaluate our framework based on a qualitative and quantitative perspective. Visually and therefore from a subjective qualitative perspective we determine that our federated T2TGAN framework has successfully modelled ABP from a single optical PPG signal alone. An example of real and generated data can be seen in Figure \ref{fig:FakeSignals} below. 

\begin{figure}[ht]
    \centering
    \includegraphics[width=0.5\textwidth]{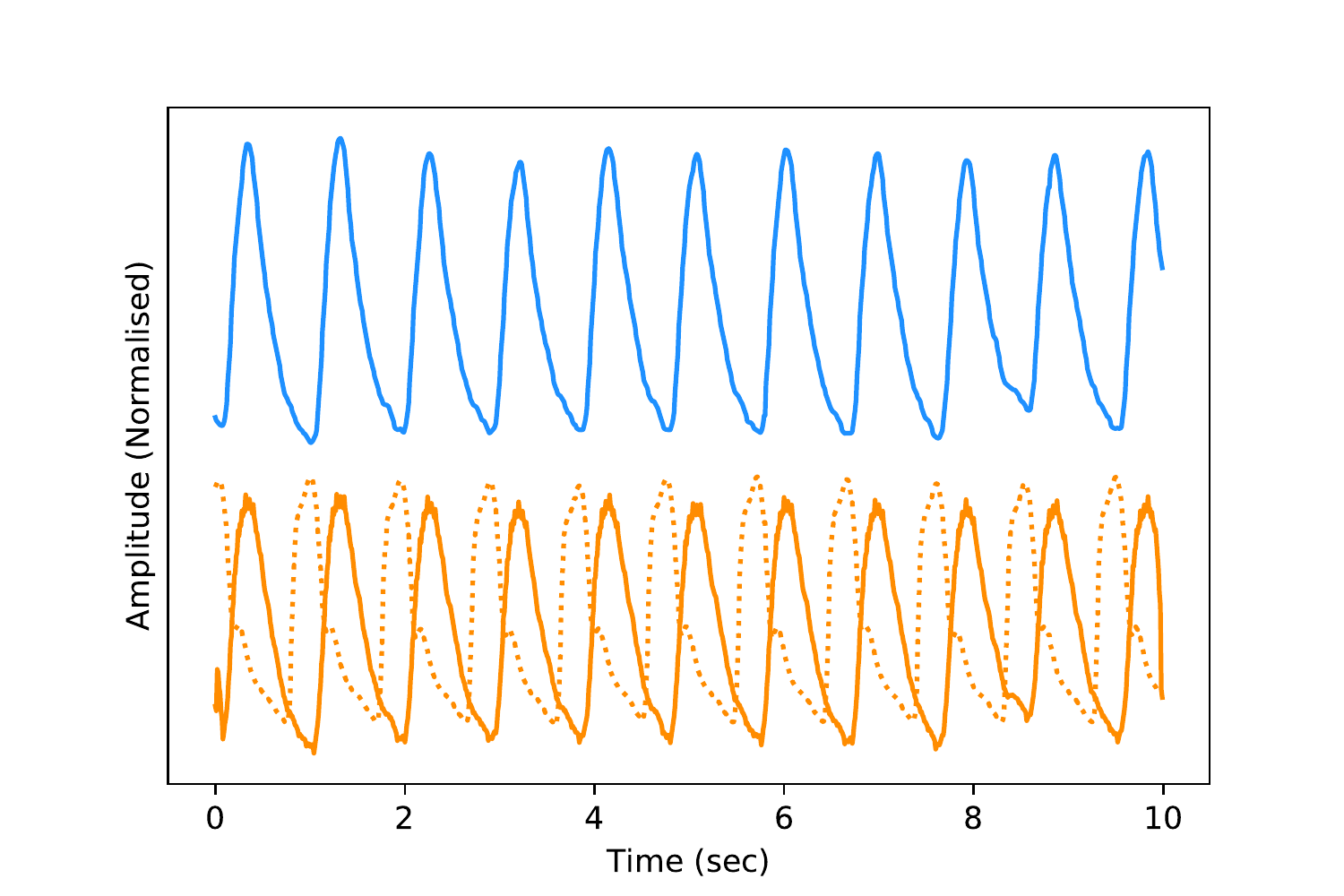}
    \caption{Example of Real PPG (top, blue) and the corresponding real ABP (bottom, dashed-orange) along with the fake ABP (bottom, orange) generated using the respective PPG. The signals are both normalised between 0 and 1 with an artificial offset on the ABP signals for visualisation purposes}
    \label{fig:FakeSignals}
\end{figure}

\begin{figure}[ht]
    \centering
    \includegraphics[width=0.45\textwidth]{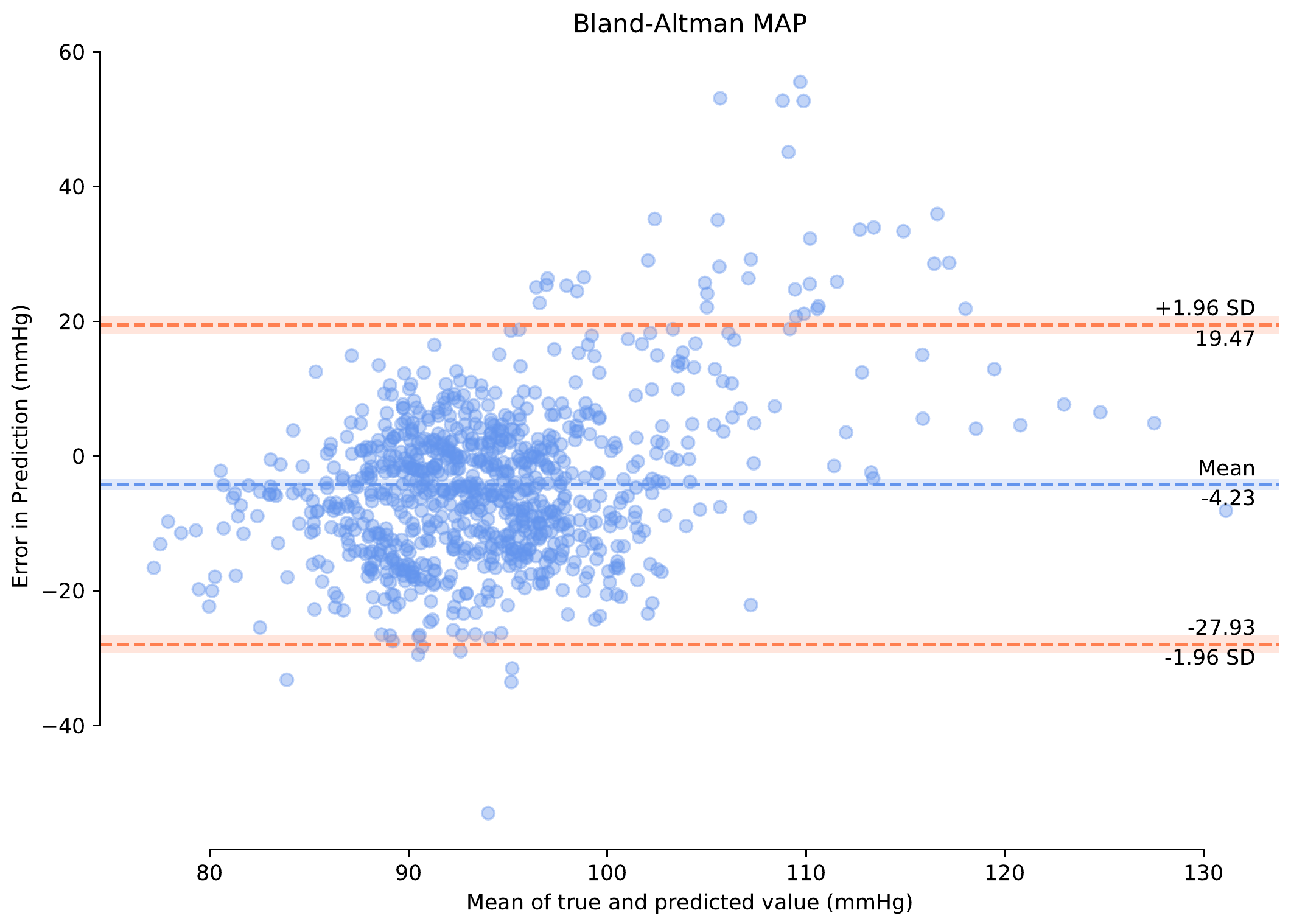}
    \includegraphics[width=0.45\textwidth]{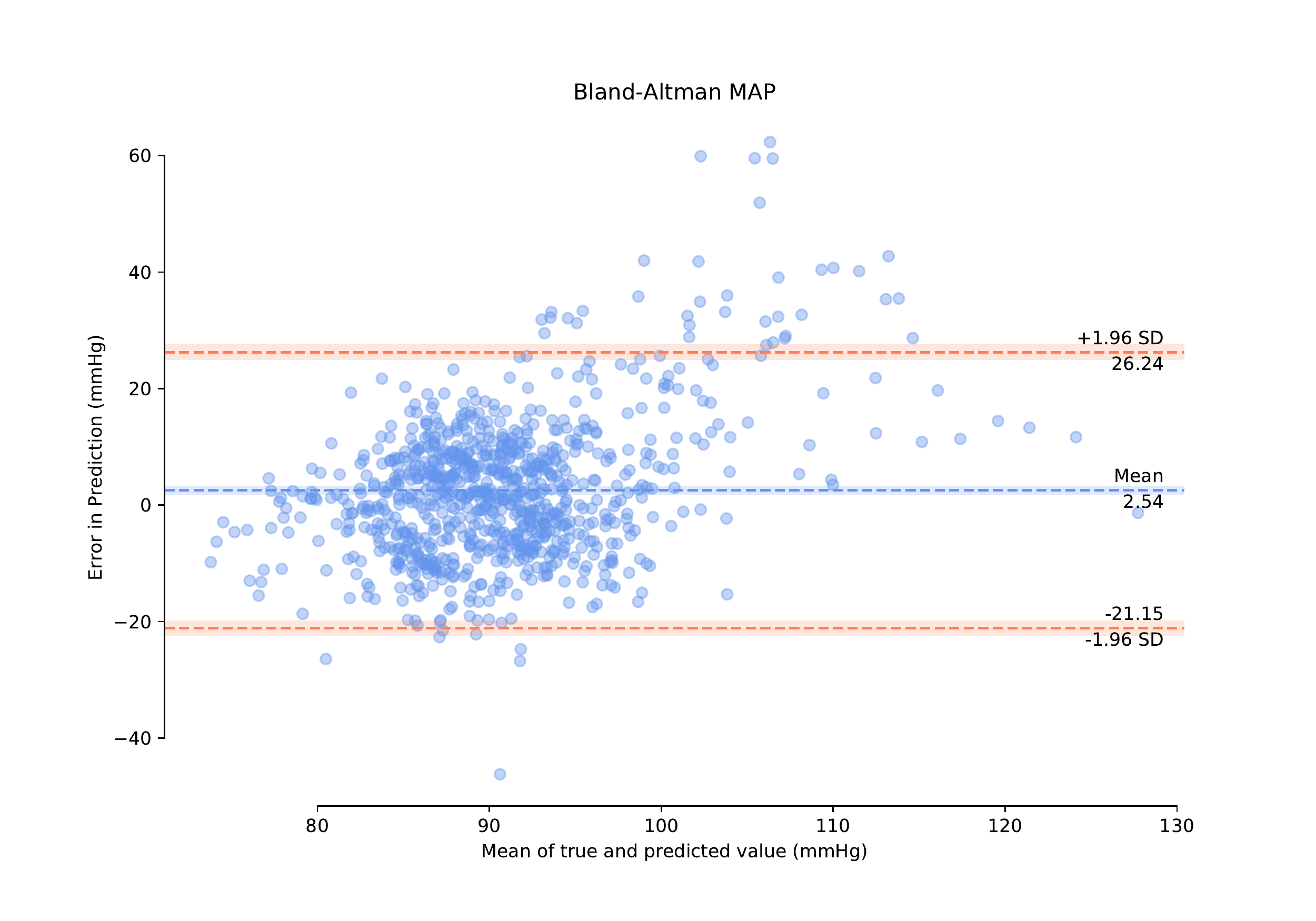}
    \caption{Bland-Altman plots of Mean Arterial Pressure on the unseen, unprocessed test data. (Left) A mean error of $-4.23  mmHg$ standard deviation of $23.7  mmHg$. (Right) Following a one-minute calibration period with a mean error of $2.54  mmHg$ standard deviation of $23.7  mmHg$.}
    \label{fig:BA_MAP}
\end{figure}

Observing the Bland-Altman plot in Figure \ref{fig:BA_MAP} (left) our framework achieved a mean \textbf{MAP error} of $-4.23  mmHg$ and a \textbf{standard deviation} of $23.7  mmHg$. We also present the Bland-Altman plots over the 149-minute period with a 1-minute calibration period that achieved a mean \textbf{MAP error of} $2.54  mmHg$ and a \textbf{standard deviation of} $23.7  mmHg$. This calibration period can prove useful in bringing the mean error to within the AAMI standards. The BA plots show the 95\% range from $\mu-1.96\sigma$ to $\mu+1.96\sigma$. The MAP range of $[26.24mmHg, -21.15mmHg]$ in Figure \ref{fig:BA_MAP} (right) demonstrate that the one-minute calibration period was successful in reducing the overall bias in the mean error.

However, a qualitative evaluation cannot alone be considered a justification of a successful framework due to the lack of a suitable objective measure. Therefore from a quantitative perspective, we compute DTW, RMSE error, and PCC of the real vs. generated continuous ABP signals. The time series similarity results on the validation and test datasets for both the federated and un-federated frameworks are displayed in Table \ref{tab:results} below. It can be seen that, as expected, the federated results are degraded slightly compared to the non-federated results. In both cases, the models perform seemly equal on the validation dataset as they do on the test dataset.

\begin{table}[ht]
\centering
 \caption{Time series similarity metrics}
 \begin{tabular}{|c|c| c c c|} 
 \hline
 Method & Dataset & DTW & RMSE & PCC \\ [0.5ex] 
 \hline\hline
 No Federated framework & Test Dataset & 56.73 & 0.19 & -0.11\\
 \hline
 No Federated framework & Validation Dataset & 55.18 & 0.23 & -0.33 \\
 \hline
 Federated framework & Test Dataset & 62.55 &0.24  &  -0.22\\
 \hline
 Federated framework & Validation Dataset & 62.15 & 0.25 & -0.34  \\ [1ex] 
 \hline
\end{tabular}
\label{tab:results}
\end{table}

\section{Discussion and Conclusion}
Here we have presented a novel decentralised learning framework for the generation of continuous ABP data and MAP estimates using a single optical sensor alone. Although our results of a mean error of $2.54  mmHg$ standard deviation of $23.7  mmHg$ do not meet the AAMI criterion it must be stated that for our test dataset we obtained a completely separate dataset and carried out no further processing on the retrieved data other than segmentation. Our framework performs deceptively well due to the real-world-ness of the test dataset and the fact that, as stated before, the physical state of patients under anesthesia contain marked changes in their cardiovascular variables (ABP and PPG in this case) in comparison to patients in the ICU (training dataset). With further work on cleaning and preprocessing the datasets, we might observe improved results, such as the results observed in \cite{ibtehaz_ppg2abp_2020}. However, in keeping with noisy real-world data, we did not implement this as part of this work.  Overall our framework lays the foundation for continuous ABP measurements on a large scale for the first time by providing a real-world example of how our models learn from small subsets of personal data and generalise seemly well to previously unseen data. All this while using a sole PPG sensor which will subsequently lead to lower-cost devices and commodtisation of such. This may be one such solution for clinicians to remotely and accurately monitor patients' cardiovascular states in their fight against CVDs.

\section*{Acknowledgments}

This work is funded by Science Foundation Ireland under grant numbers 17/RC-PhD/3482 and SFI/12/RC/2289\_P2. We also gratefully acknowledge the support of NVIDIA Corporation with the donation of the Titan Xp used for this research.

\bibliographystyle{IEEEtran}
\bibliography{bibliography/references}

% Generated by IEEEtran.bst, version: 1.14 (2015/08/26)
\begin{thebibliography}{10}
\providecommand{\url}[1]{#1}
\csname url@samestyle\endcsname
\providecommand{\newblock}{\relax}
\providecommand{\bibinfo}[2]{#2}
\providecommand{\BIBentrySTDinterwordspacing}{\spaceskip=0pt\relax}
\providecommand{\BIBentryALTinterwordstretchfactor}{4}
\providecommand{\BIBentryALTinterwordspacing}{\spaceskip=\fontdimen2\font plus
\BIBentryALTinterwordstretchfactor\fontdimen3\font minus
  \fontdimen4\font\relax}
\providecommand{\BIBforeignlanguage}[2]{{%
\expandafter\ifx\csname l@#1\endcsname\relax
\typeout{** WARNING: IEEEtran.bst: No hyphenation pattern has been}%
\typeout{** loaded for the language `#1'. Using the pattern for}%
\typeout{** the default language instead.}%
\else
\language=\csname l@#1\endcsname
\fi
#2}}
\providecommand{\BIBdecl}{\relax}
\BIBdecl

\bibitem{WHO_2020}
\BIBentryALTinterwordspacing
``The top 10 causes of death,'' World Health Organisation, 2020. [Online].
  Available:
  \url{https://www.who.int/news-room/fact-sheets/detail/the-top-10-causes-of-death}
\BIBentrySTDinterwordspacing

\bibitem{padilla_2006}
J.~M. {Padilla}, E.~J. {Berjano}, J.~{Saiz}, L.~{Facila}, P.~{Diaz}, and
  S.~{Merce}, ``Assessment of relationships between blood pressure, pulse wave
  velocity and digital volume pulse,'' in \emph{2006 Computers in Cardiology},
  2006, pp. 893--896.

\bibitem{wong_evaluation_2009}
\BIBentryALTinterwordspacing
M.~Y.-M. Wong, C.~C.-Y. Poon, and Y.-T. Zhang,
  ``\BIBforeignlanguage{english}{An {Evaluation} of the {Cuffless} {Blood}
  {Pressure} {Estimation} {Based} on {Pulse} {Transit} {Time} {Technique}: a
  {Half} {Year} {Study} on {Normotensive} {Subjects}},''
  \emph{\BIBforeignlanguage{english}{Cardiovascular Engineering}}, vol.~9,
  no.~1, pp. 32--38, Mar. 2009. [Online]. Available:
  \url{http://link.springer.com/10.1007/s10558-009-9070-7}
\BIBentrySTDinterwordspacing

\bibitem{brophy2020optimised}
E.~Brophy, W.~Muehlhausen, A.~F. Smeaton, and T.~E. Ward, ``Optimised
  convolutional neural networks for heart rate estimation and human activity
  recognition in wrist worn sensing applications,'' 2020.

\bibitem{teng_continuous_2003}
\BIBentryALTinterwordspacing
X.~Teng and Y.~Zhang, ``\BIBforeignlanguage{english}{Continuous and noninvasive
  estimation of arterial blood pressure using a photoplethysmographic
  approach},'' in \emph{\BIBforeignlanguage{english}{Proceedings of the 25th
  {Annual} {International} {Conference} of the {IEEE} {Engineering} in
  {Medicine} and {Biology} {Society} ({IEEE} {Cat}. {No}.{03CH37439})}}.\hskip
  1em plus 0.5em minus 0.4em\relax Cancun, Mexico: IEEE, 2003, pp. 3153--3156.
  [Online]. Available: \url{http://ieeexplore.ieee.org/document/1280811/}
\BIBentrySTDinterwordspacing

\bibitem{Lin_2021}
\BIBentryALTinterwordspacing
W.-H. Lin, F.~Chen, Y.~Geng, N.~Ji, P.~Fang, and G.~Li, ``Towards accurate
  estimation of cuffless and continuous blood pressure using multi-order
  derivative and multivariate photoplethysmogram features,'' \emph{Biomedical
  Signal Processing and Control}, vol.~63, p. 102198, 2021. [Online].
  Available:
  \url{https://www.sciencedirect.com/science/article/pii/S1746809420303359}
\BIBentrySTDinterwordspacing

\bibitem{slapnicar_blood_2019}
\BIBentryALTinterwordspacing
G.~Slapni\v{c}ar, N.~Mlakar, and M.~Lu\v{s}trek, ``Blood pressure estimation
  from photoplethysmogram using a spectro-temporal deep neural network,''
  \emph{Sensors}, vol.~19, no.~15, 2019. [Online]. Available:
  \url{https://www.mdpi.com/1424-8220/19/15/3420}
\BIBentrySTDinterwordspacing

\bibitem{Hajj_2020}
C.~{El Hajj} and P.~A. {Kyriacou}, ``Cuffless and continuous blood pressure
  estimation from ppg signals using recurrent neural networks,'' in \emph{2020
  42nd Annual International Conference of the IEEE Engineering in Medicine
  Biology Society (EMBC)}, 2020, pp. 4269--4272.

\bibitem{Nath_2020}
R.~K. {Nath} and H.~{Thapliyal}, ``Ppg based continuous blood pressure
  monitoring framework for smart home environment,'' in \emph{2020 IEEE 6th
  World Forum on Internet of Things (WF-IoT)}, 2020, pp. 1--6.

\bibitem{Schlesinger_2020}
O.~{Schlesinger}, N.~{Vigderhouse}, D.~{Eytan}, and Y.~{Moshe}, ``Blood
  pressure estimation from ppg signals using convolutional neural networks and
  siamese network,'' in \emph{ICASSP 2020 - 2020 IEEE International Conference
  on Acoustics, Speech and Signal Processing (ICASSP)}, 2020, pp. 1135--1139.

\bibitem{Panwar_2020}
M.~{Panwar}, A.~{Gautam}, D.~{Biswas}, and A.~{Acharyya}, ``Pp-net: A deep
  learning framework for ppg-based blood pressure and heart rate estimation,''
  \emph{IEEE Sensors Journal}, vol.~20, no.~17, pp. 10\,000--10\,011, 2020.

\bibitem{ibtehaz_ppg2abp_2020}
\BIBentryALTinterwordspacing
N.~Ibtehaz and M.~S. Rahman, ``\BIBforeignlanguage{english}{{PPG2ABP}:
  {Translating} {Photoplethysmogram} ({PPG}) {Signals} to {Arterial} {Blood}
  {Pressure} ({ABP}) {Waveforms} using {Fully} {Convolutional} {Neural}
  {Networks}},'' \emph{\BIBforeignlanguage{english}{arXiv:2005.01669 [cs,
  eess]}}, May 2020, arXiv: 2005.01669. [Online]. Available:
  \url{http://arxiv.org/abs/2005.01669}
\BIBentrySTDinterwordspacing

\bibitem{Mena_2020}
\BIBentryALTinterwordspacing
L.~J. Mena, V.~G. F{\'e}lix, R.~Ostos, A.~J. Gonz{\'a}lez,
  R.~Mart{\'i}nez-Pel{\'a}ez, J.~D. Melgarejo, and G.~E. Maestre, ``Mobile
  personal health care system for noninvasive, pervasive, and continuous blood
  pressure monitoring: Development and usability study,'' \emph{JMIR Mhealth
  Uhealth}, vol.~8, no.~7, p. e18012, Jul 2020. [Online]. Available:
  \url{https://mhealth.jmir.org/2020/7/e18012}
\BIBentrySTDinterwordspacing

\bibitem{sarkar_cardiogan_2020}
\BIBentryALTinterwordspacing
P.~Sarkar and A.~Etemad, ``\BIBforeignlanguage{english}{{CardioGAN}:
  {Attentive} {Generative} {Adversarial} {Network} with {Dual} {Discriminators}
  for {Synthesis} of {ECG} from {PPG}},''
  \emph{\BIBforeignlanguage{english}{arXiv:2010.00104 [cs, eess]}}, Dec. 2020,
  arXiv: 2010.00104. [Online]. Available: \url{http://arxiv.org/abs/2010.00104}
\BIBentrySTDinterwordspacing

\bibitem{Google_FL_2017}
\BIBentryALTinterwordspacing
``Federated learning: Collaborative machine learning without centralized
  training data,'' Google AI, 2017. [Online]. Available:
  \url{https://ai.googleblog.com/2017/04/federated-learning-collaborative.html}
\BIBentrySTDinterwordspacing

\bibitem{rasouli2020fedgan}
M.~Rasouli, T.~Sun, and R.~Rajagopal, ``Fedgan: Federated generative
  adversarial networks for distributed data,'' 2020.

\bibitem{zhu2020CycleGAN}
J.-Y. Zhu, T.~Park, P.~Isola, and A.~A. Efros, ``Unpaired image-to-image
  translation using cycle-consistent adversarial networks,'' 2020.

\bibitem{kachuee_cuff-less_2015}
\BIBentryALTinterwordspacing
M.~Kachuee, M.~M. Kiani, H.~Mohammadzade, and M.~Shabany,
  ``\BIBforeignlanguage{english}{Cuff-less high-accuracy calibration-free blood
  pressure estimation using pulse transit time},'' in
  \emph{\BIBforeignlanguage{english}{2015 {IEEE} {International} {Symposium} on
  {Circuits} and {Systems} ({ISCAS})}}.\hskip 1em plus 0.5em minus 0.4em\relax
  Lisbon, Portugal: IEEE, May 2015, pp. 1006--1009. [Online]. Available:
  \url{http://ieeexplore.ieee.org/document/7168806/}
\BIBentrySTDinterwordspacing

\bibitem{kachuee_cuffless_2017}
M.~{Kachuee}, M.~M. {Kiani}, H.~{Mohammadzade}, and M.~{Shabany}, ``Cuffless
  blood pressure estimation algorithms for continuous health-care monitoring,''
  \emph{IEEE Transactions on Biomedical Engineering}, vol.~64, no.~4, pp.
  859--869, 2017.

\bibitem{PhysioNet}
A.~L. Goldberger, L.~A.~N. Amaral, L.~Glass, J.~M. Hausdorff, P.~C. Ivanov,
  R.~G. Mark, J.~E. Mietus, G.~B. Moody, C.-K. Peng, and H.~E. Stanley,
  ``{PhysioBank, PhysioToolkit, and PhysioNet}: Components of a new research
  resource for complex physiologic signals,'' \emph{Circulation}, vol. 101,
  no.~23, pp. e215--e220, 2000 (June 13), circulation Electronic Pages:
  http://circ.ahajournals.org/content/101/23/e215.full PMID:1085218; doi:
  10.1161/01.CIR.101.23.e215.

\bibitem{liu_university_2012}
\BIBentryALTinterwordspacing
D.~Liu, M.~Görges, and S.~A. Jenkins,
  ``\BIBforeignlanguage{english}{University of {Queensland} {Vital} {Signs}
  {Dataset}: {Development} of an {Accessible} {Repository} of {Anesthesia}
  {Patient} {Monitoring} {Data} for {Research}},''
  \emph{\BIBforeignlanguage{english}{Anesthesia \& Analgesia}}, vol. 114,
  no.~3, pp. 584--589, Mar. 2012. [Online]. Available:
  \url{http://journals.lww.com/00000539-201203000-00015}
\BIBentrySTDinterwordspacing

\bibitem{MAP_2011}
\BIBentryALTinterwordspacing
L.~Bonsall, ``Calculating the mean arterial pressure (map),'' Lippincott
  NursingCenter, 2011. [Online]. Available:
  \url{https://www.nursingcenter.com/ncblog/december-2011/calculating-the-map}
\BIBentrySTDinterwordspacing

\bibitem{Giavarina_2015}
D.~Giavarina, ``Understanding bland altman analysis,'' \emph{Croatian Society
  of Medical Biochemistry and Laboratory Medicine}, vol.~25, pp. 141--51, 6
  2015.

\bibitem{AAMI_1987}
\BIBentryALTinterwordspacing
A.~for the Advancement~of Medical~Instrumentation, ``American national
  standards for electronic or automated sphygmomanometers,'' \emph{ANSI/AAMI SP
  10-1987}, 1987. [Online]. Available:
  \url{https://ci.nii.ac.jp/naid/10024828510/en/}
\BIBentrySTDinterwordspacing

\end{thebibliography}

\end{document}